\documentclass[final]{cvpr}

\usepackage{times}
\usepackage{epsfig}
\usepackage{graphicx}
\usepackage{amsmath}
\usepackage{amssymb}
\usepackage{multirow}
\usepackage[pagebackref=true,breaklinks=true,colorlinks,urlcolor=blue,bookmarks=false]{hyperref}

\def\cvprPaperID{****} % *** Enter the CVPR Paper ID here
\def\confYear{CVPR 2021}

\begin{document}
	
	%%%%%%%%% TITLE
	\title{Cascade Bagging for Accuracy Prediction with  Few Training Samples}
	
	\author{Ruyi Zhang\textsuperscript{1},Ziwei Yang\textsuperscript{1},  Zhi Yang\textsuperscript{1}, Xubo Yang\textsuperscript{1}, Lei Wang\textsuperscript{3} and Zheyang Li\textsuperscript{1,2}\\
		\textsuperscript{1}Hikvision Research Institute, \textsuperscript{2}Zhejiang University \\
		\textsuperscript{3}University of Science and Technology of China\\
		{\tt\small \{zhangruyi5, yangziwei5,  yangzhi13, yangxubo, lizheyang\}@hikvision.com\tt\small}\\ \tt\small wangl26@mail.ustc.edu.cn
	}
	\maketitle

	%%%%%%%%% ABSTRACT
	\begin{abstract}
		%为什么预测精度
Accuracy predictor is trained to predict the validation accuracy of an network from its architecture encoding. It can effectively assist in designing networks and improving Neural Architecture Search(NAS) efficiency. However, a high-performance predictor depends on adequate trainning samples, which requires unaffordable computation overhead.  To alleviate this problem, we propose a novel framework to train an accuracy predictor under few training samples.  The framework consists of data augmentation methods and an ensemble learning algorithm.   The data augmentation methods calibrate  weak labels and inject noise to feature space. The ensemble learning algorithm, termed cascade bagging,  trains two-level models by sampling data and features. In the end, the advantages of above methods are proved in the Performance Prediciton Track of   CVPR2021 1st Lightweight NAS Challenge. Our code is made public at: \url{https://github.com/dlongry/Solution-to-CVPR2021-NAS-Track2}
		
	\end{abstract}
	
	%这里画一个多阶段bagging的图
	\begin{figure}[h]
		\begin{center}
			\includegraphics[width=1\linewidth]{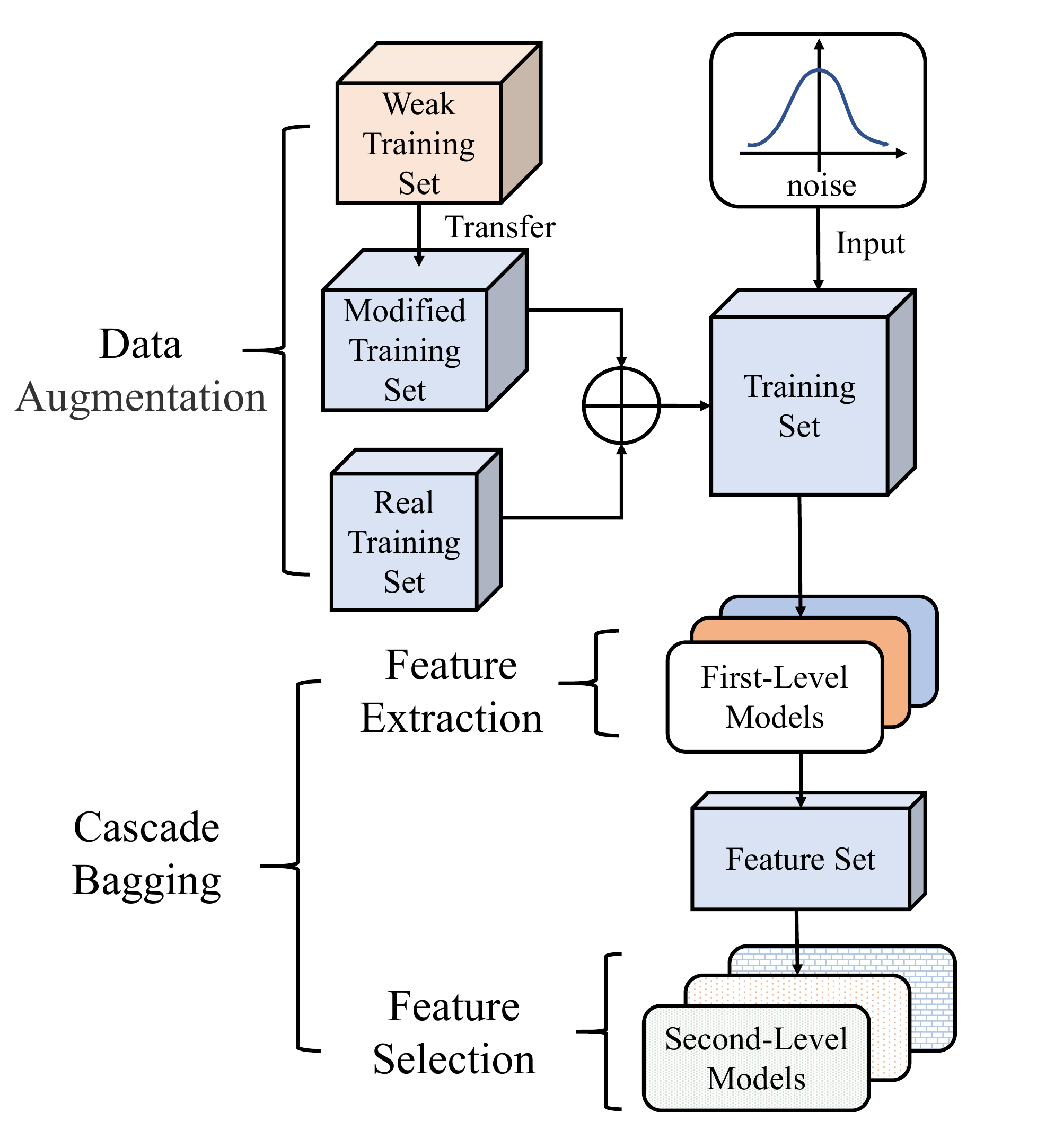}
		\end{center}
		\caption{The overview framework of our method,consisting of data augmentation and cascade bagging algorithm.}
		\label{all_system_scheme}
	\end{figure}
	
	%%%%%%%%% BODY TEXT
	\section{Introduction}
	Neural architecture search advances beyond the state-of-the-art in various computer vision tasks. However, it often trains and evaluates a large number of architectures, causing tremendous computation costs. For instance, Zoph et al.\cite{DBLP:journals/corr/ZophVSL17} spends more than 1800 GPU days and Real et al.\cite{DBLP:journals/corr/abs-1802-01548} uses 450 GPUs for 7 days to train and evaluate the models. Thus, how to estimate the performance of a neural architecture in a fast and accurate way is vital for addressing the computational challenge of NAS.

	%加上预测精度的定义
	%先确认问题定义：
	% 参考文献在A Surgery of the Neural Architecture Evaluators
	Predictor-based evaluation strategy is one of the mainstream  methods to reduce the computation overhead. It  takes the architecture description as inputs and outputs a predicted performance score. Two factors are crucial to the predictor fitness: 1) embedding space; 2) regression model. To embed neural architectures into a continuous space and get a meaningful embedding space, many studies have proposed different encoders, such as sequence-based  methods\cite{DBLP:journals/corr/abs-1808-07233,DBLP:journals/corr/abs-1712-00559,DBLP:journals/corr/abs-1712-03351,DBLP:journals/corr/abs-1903-11059}  and graph-based methods\cite{DBLP:journals/corr/abs-2004-08423,dudziak2020brp,DBLP:journals/corr/abs-2004-01899}.  Several different regression models have been utilized to estimate the accuracy, including gradient boosting  decision  tree\cite{2020Neural}, gaussian process\cite{2020GP} and graph convolution network\cite{dudziak2020brp,DBLP:journals/corr/abs-2004-08423}.
	
	In this paper, to reduce the computation burden of predictor-based evaluation strategy, we propose a novel framework to train accuracy predictor under few training samples.  The framework consists of data augmentation with limited budget and ensemble learning for promoting generalization. In the end, the effectiveness of the framework was verified in the Performance Prediction Track of CVPR2021 1st Lightweight NAS Challenge.

	\begin{figure*}[h]
		\begin{center}
			\includegraphics[width=0.95\linewidth]{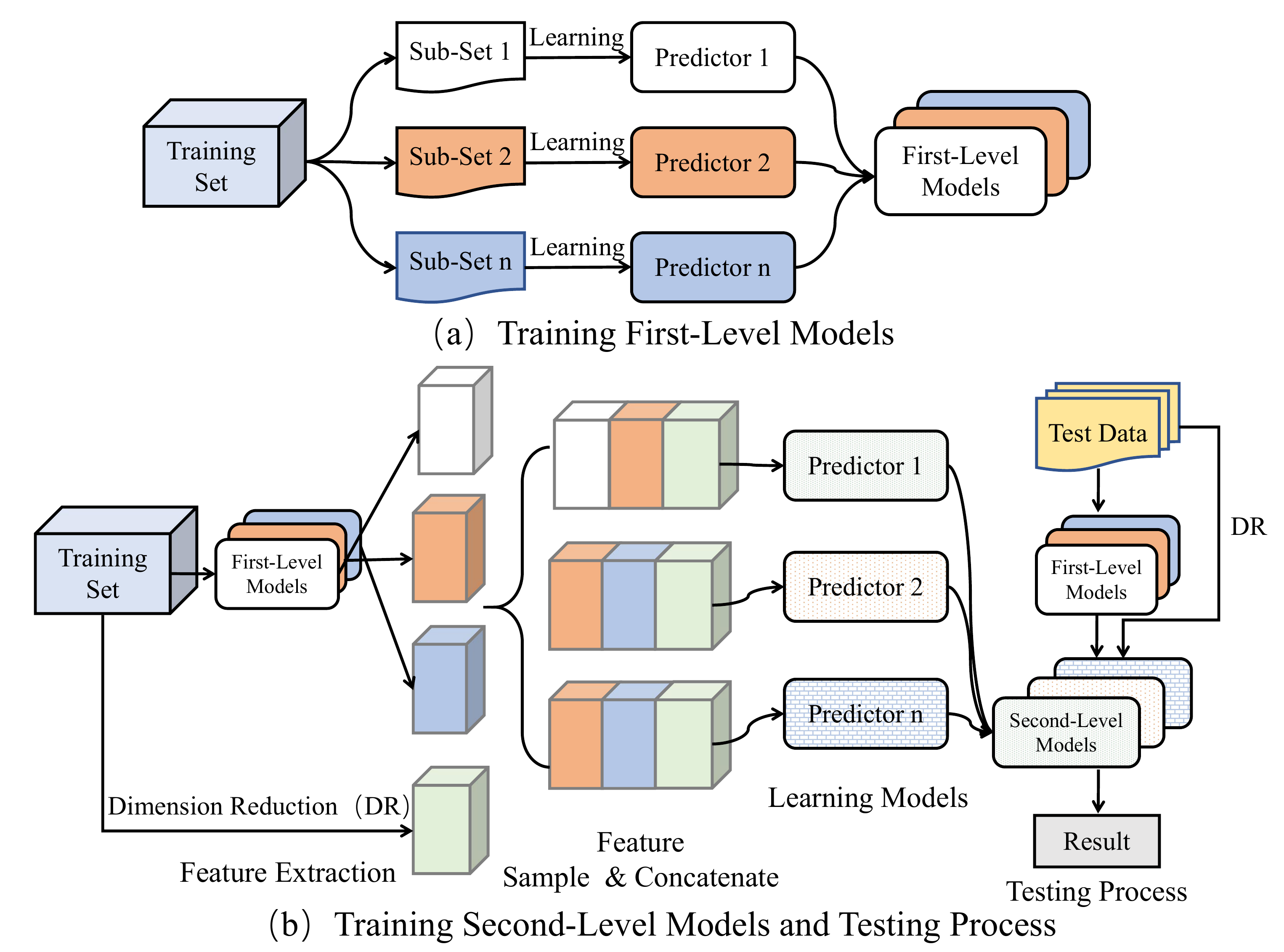}
		\end{center}
		\caption{The training framework  of the cascade bagging. (a) The training diagram of the first-level models. (b) The training diagram of the second-level models and testing process.}
		\label{whole_framework}
	\end{figure*}
	
	\section{Method}
	For reducing the cost of NAS, we propose a novel framework to train an accuracy predictor under few training samples. To alleviate the high risk of overfitting caused by the lack of training data, the framework consists of data augmentation methods and an ensemble method termed cascade bagging.
	
	\subsection{Data Augmentation}
	 We adopt  weak label calibration and gussian noise additon to augment dataset under limited computation overhead.
	
	\subsubsection{Label Calibration} 
	
	Given two datasets $S_r$ and $S_{w}$. $S_r$ only has few architectures with ground-truth labels $Y_r$.  $S_{w}$ has more architectures with weak labels $Y_w$. Weak labels are generated by insufficient training under limited overhead. Here, we calibrate the weak labels $Y_w$ and make it close to the ground-truth label. 
	
	Two valid information are chosen to calibrate the weak labels.  One is the difference between the mean values of  $Y_r$ and $Y_w$. The other  is the difference between the weak labels and the prediction results. Specifically, we divide the samples of $S_w$  into $N$ groups according to accuracy.  We assume that the weak labels and ground truth labels have the similar biases. Based on the assumption, i-th weak label $Y_{w_i}$ in a group can be calibrated with a function as follows:
	\begin{equation}
	Y_{w_i}^{'} = Y_{w_i}+(E_r-E_w)+b 
	\end{equation} 
	$E_r$ and $E_w$ are the mean values of $Y_r$ and $Y_w$, respectively.  $b$ means modified bias, which is a hyper-parameter. 
	
	 Referring to the ideas of semi-supervised algorithm \cite{2006Semi,2019Self}, we  utilize the  prediction results to correct weak labels with the following equation:
	\begin{equation}
	Y_{w_i}^{'} = \alpha * Y_{w_i}+(1-\alpha) * Y_{P_i}
	\end{equation} 
	$Y_{P_i}$ is the prediction result of i-th architecture. $\alpha$ denotes a linear coefficient hyper-parameter. Note that, since a new predictor can be  updated with modified labels. The new $Y_{P_i}$ can be generated. The calibration process above can be made repeatly. We find that the prediction results from multiple calibration processes  benefit for final predictor performance.
	
	\subsubsection{Noise addition } 
	Many studies\cite{bishop1995training,2019Self} have reported that adding noise to smooth feature space can improve model generalization ability. In this work,  we utilize noise to enlarge training set with negligible cost. we firstly sample a data (x,y) from training set. We then inject  Gaussian noise to slightly perturb the  feature x to get new feature x'. Lastly, give the feature x' the same label y as x, we can obtain new training data (x',y). 
	
	\subsection{Cascade Bagging}
	 Traditional bagging algorithm has ability to promote models' generalization\cite{2019The}. Utilizing traditional bagging algorithm to train a predictor has two steps. Firstly, numerous sub training sets are created by sampling data from training set with repleacement. Secondly, a series of weak predictors are trained with the sub training sets, repectively. In testing phase, the final prediction result is mean value of the preditors' outputs.   Traditional bagging algorithm ignores the fact that different predictors should make different contribution to final prediction result.  For automatically calculating contributions of different predictors, we propose a cascade bagging algorithm in our framework.  
	
	As shown in Figure 2, we firstly train numerous first-level models with the sub training sets, which are  the same as the traditional bagging algorithm. Secondly, we train second-level models using different combinations of features.  As shown in the Figure 2(b), the combination features  are  consisted of two parts: 1)the first part  features are sampled from the outputs of first-level models; 2)the second part features are architecture encoding after PCA dimension reducing. 
	
	 In testing phase, we calculate the outputs of  first-level models and  reduce the dimension of architecture encoding.  We concatenate the outputs and the  dimension reduction features prior to inputting them into second-level models.  Lastly, the second-level outputs are averaged as the final prediction result.
	
	\section{Experiment}
	In this work, we proposed a novel framework to train an accuracy predictor with few training samples. This section reports the effectiveness of the framework by various experiments. 
	
	\subsection{Dataset}
	The dataset are supplied by CVPR2021 workshop NAS challenge. It contains 231 training samples. 200 samples  have weak labels, which are obtained by training models with insufficient epoches.  Other 31 samples have ground truth labels obtained by sufficient training. 
	
	\subsection{Architectures Encoder}
	The  architectures of the competition dataset are sampled from the Mobilenet-like search space, where 16 blocks are searchable. The 16 blocks are connected to each other in sequence.  The choices of each block range from [1,6] which means 6 (three choices of kernel size, two choices of expansion rate) different operations. 
	
	%基于上述空间，编码有黑箱编码和白箱编码两种形式。黑箱编码一般索引编码，即一个网络用16维特征来表示，第i维表示第i个block选择的操作数的索引。 另一种使白箱编码，它表示一个网络用2x16维特征表示，第i维的两维向量分别表示第i个block的kernel size和expansion的大小
	Based on the above space, we design a black-box and a white-box architecture encoding. Black-box encodes a network with  16-dimensional features, and the i-th dimension denotes the index of the operation  of the i-th block. White-box encodes the network as a $2\times16$-dimensional tensor. Each block is represented by a two-dimensional vector which denotes its kernel size and expansion ratio.
	
	\subsection{Ablation Study}
	\begin{table}[h]
		\begin{center}
			\begin{tabular}{cccccc}
				\hline
				method  & encode & g  & bias & noise & RMSE \\
				\hline
				B1 &  Black    &  -     &  -   &  w/o    &  0.252    \\
				B2 &  Black    &  1     &  2.2 &  w/o     & 0.228     \\
				W1 &  White    &  1     &  2.2 &  w/o     & 0.212     \\
				W2 &  White    &  3     &  2.24,2.2,2.22   & w/o     &  0.204    \\
				W3 &  White    &  3     &  2.24,2.2,2.22   & w     &  0.201    \\			
				\hline
			\end{tabular}
		\end{center}
		\caption{Ablation study of label calibration with eqn(1)  and noise addition. "encode" means encoding type of architectures. "g" and "bias" means group number and value of  $(E_r-E_w)+b$ in eqn(1), respectively. "noise" indicates whether noise is added to feature space. }
	\end{table}
	The experiments in Table1  prove the validity of  calibration labels with adding proper bias. These experiments base on Xgboost model. Method B1 only uses 31 samples with ground truth label.  Compared to B1, B2 uses more 200 calibrated samples. The result of B2 presents huge improvement over B1, which illustrates adding proper bias is effective for label calibration. The performance of W3 is better than W2, indicating the validity of noise addition for generating new data. In addition, the results of B2 and W1 highlight that white-box is a better architecture encoder than black-box. 
	
	\begin{table}[h]
		\begin{center}
			\begin{tabular}{cccc}
				\hline
				method & models for calibraition  &  iter  & RMSE \\
				\hline
				S1 &   -             &  -        &       0.1857   \\
				S2 &   SVM           &  1        &       0.1849   \\
				S3 &   SVM           &  3        &       0.1846   \\
				S4 &  SVM,KNN        &  3,1      &       0.1849   \\	
				S5 &  SVM,Xgboost,   &  3,1      &       0.1829   \\
				\hline
			\end{tabular}
		\end{center}
		\caption{Ablation study of  label calibration with eqn(2). "iter" denotes number of calibration times. In S4, it means the calibration times of SVM and KNN are 3 and 1 respectively.}
	\end{table}
	
	The experiments in Table2  verify the effictiveness  of weighting sum the prediction results for calibrating labels. The experiments base on SVM model and use traditional bagging algorithm, which train 100 models with random data sampling.  Because of the above changes, the RMSE decreases from 0.201 to 0.1857.  S2 and S3 calibrate the weak labels using Eqn(2) with 1 group and 3 group, respectively. The results show that  prediction results are effective information to calibrate labels and multiple calibration processes also have positive effect on the final result. In addition, S5 utilizes  prediction result from SVM and xgboost to correct the weak labels. Its results prove that using more models may further improve performance. However, S4 shows that this improvement does not persist for all models. The reason may be that label noise is introduced from  poor prediction results.
	
	%表2的实验全都使用了bagging，放回采样训练了100个模型，并对所有结果取均值输出。，并进行了tabel1的数据增强操作。table1中的xgboost基模型换成了SVM。相比于tabel1的实验四，经过这些操作后rmse变为0.1857变成了这次实验的baseline，证明了bagging的有效性。同时对比实验2和实验三，发现多次迭代，知道收敛的模型修正也是有效的。使用其他模型进行修正也可以提高模型性能。但是存在过拟合风险。
	
	\begin{table}[h]
		\begin{center}
			\begin{tabular}{cc}
				\hline
				 method            &  RMSE\\
				\hline
				 Bagging               &  0.1829   \\       
				 Cascade bagging      &  0.1808   \\
				\hline
			\end{tabular}
		\end{center}
		\caption{Ablation study of cascade bagging}
	The experiments in Table3  prove validity of cascade bagging. Besides training 100 first-level models, the cascade bagging trains  more 100 second-level models. The second-level models enble to  automatically calculate first-level models' contribution. The results in the table show that  cascade bagging can further enhance the  predictor  performance.
		
	\end{table}
	\section{Conclusion}
	This work provides a novel framework to train an accuracy predictor under few training samples.  The framework consists of data augmentation methods and a cascade bagging algorithm.  By conducting experiments on different combinations of the training methods, we find that the prediction results and the mean value of ground truth label  are valid information for label calibration. We also observe that injecting noises to feature space is able to generate effective data. In addition, our cascade bagging algorithm can automatically adjust the contribution of different preditors, which outperforms traditional bagging algorithm. Finally, combined with the above methods, our framework ranks the 3rd place in the Performance Prediction Track of CVPR2021 1st Lightweight NAS Challenge.

	{\small
		\bibliographystyle{ieee_fullname}
		\bibliography{egbib}

\begin{thebibliography}{10}\itemsep=-1pt

\bibitem{bishop1995training}
Chris~M Bishop.
\newblock Training with noise is equivalent to tikhonov regularization.
\newblock {\em Neural computation}, 7(1):108--116, 1995.

\bibitem{DBLP:journals/corr/abs-2004-08423}
Xin Chen, Lingxi Xie, Jun Wu, Longhui Wei, Yuhui Xu, and Qi Tian.
\newblock Fitting the search space of weight-sharing {NAS} with graph
  convolutional networks.
\newblock {\em CoRR}, abs/2004.08423, 2020.

\bibitem{DBLP:journals/corr/abs-1712-03351}
Boyang Deng, Junjie Yan, and Dahua Lin.
\newblock Peephole: Predicting network performance before training.
\newblock {\em CoRR}, abs/1712.03351, 2017.

\bibitem{dudziak2020brp}
{\L}ukasz Dudziak, Thomas Chau, Mohamed~S Abdelfattah, Royson Lee, Hyeji Kim,
  and Nicholas~D Lane.
\newblock Brp-nas: Prediction-based nas using gcns.
\newblock {\em arXiv preprint arXiv:2007.08668}, 2020.

\bibitem{2019The}
B. Ghojogh and M. Crowley.
\newblock The theory behind overfitting, cross validation, regularization,
  bagging, and boosting: Tutorial.
\newblock 2019.

\bibitem{2006Semi}
Mfa Hady and F.Schwenker.
\newblock Semi-supervised learning.
\newblock 2006.

\bibitem{2020GP}
Z. Li, T. Xi, J. Deng, G. Zhang, and R. He.
\newblock Gp-nas: Gaussian process based neural architecture search.
\newblock In {\em 2020 IEEE/CVF Conference on Computer Vision and Pattern
  Recognition (CVPR)}, 2020.

\bibitem{DBLP:journals/corr/abs-1712-00559}
Chenxi Liu, Barret Zoph, Jonathon Shlens, Wei Hua, Li{-}Jia Li, Li Fei{-}Fei,
  Alan~L. Yuille, Jonathan Huang, and Kevin Murphy.
\newblock Progressive neural architecture search.
\newblock {\em CoRR}, abs/1712.00559, 2017.

\bibitem{2020Neural}
R. Luo, X. Tan, R. Wang, T. Qin, E. Chen, and T.~Y. Liu.
\newblock Neural architecture search with gbdt.
\newblock 2020.

\bibitem{DBLP:journals/corr/abs-1808-07233}
Renqian Luo, Fei Tian, Tao Qin, and Tie{-}Yan Liu.
\newblock Neural architecture optimization.
\newblock {\em CoRR}, abs/1808.07233, 2018.

\bibitem{DBLP:journals/corr/abs-2004-01899}
Xuefei Ning, Yin Zheng, Tianchen Zhao, Yu Wang, and Huazhong Yang.
\newblock A generic graph-based neural architecture encoding scheme for
  predictor-based {NAS}.
\newblock {\em CoRR}, abs/2004.01899, 2020.

\bibitem{DBLP:journals/corr/abs-1802-01548}
Esteban Real, Alok Aggarwal, Yanping Huang, and Quoc~V. Le.
\newblock Regularized evolution for image classifier architecture search.
\newblock {\em CoRR}, abs/1802.01548, 2018.

\bibitem{DBLP:journals/corr/abs-1903-11059}
Linnan Wang, Yiyang Zhao, Yuu Jinnai, Yuandong Tian, and Rodrigo Fonseca.
\newblock Alphax: exploring neural architectures with deep neural networks and
  monte carlo tree search.
\newblock {\em CoRR}, abs/1903.11059, 2019.

\bibitem{2019Self}
Q. Xie, M.~T. Luong, E. Hovy, and Q.~V. Le.
\newblock Self-training with noisy student improves imagenet classification.
\newblock 2019.

\bibitem{DBLP:journals/corr/ZophVSL17}
Barret Zoph, Vijay Vasudevan, Jonathon Shlens, and Quoc~V. Le.
\newblock Learning transferable architectures for scalable image recognition.
\newblock {\em CoRR}, abs/1707.07012, 2017.

\end{thebibliography}
	}
	
\end{document}